\documentclass[12pt]{article}

\usepackage{sbc-template}

\usepackage{graphicx,url}

\usepackage[brazil]{babel}   
\usepackage[utf8]{inputenc}  
\usepackage[T1]{fontenc}
\usepackage{enumitem}

\title{Teleoperando Robôs Pioneer Utilizando Android}

\author{Eduardo Gouveia Pinheiro\inst{1}, Túlio Casagrande Alberto\inst{1}}

\address{Departamento de Computação -- Universidade Federal de São Carlos
  (UFSCar)\\
  Sorocaba -- SP -- Brasil
  \email{\{edu.g.pinheiro, tuliocasagrande\}@gmail.com}
}

\begin{document} 
\maketitle

\begin{abstract}
This paper presents an application with ROS, Aria and RosAria to control a ModelSim simulated Pioneer 3-DX robot. The navigation applies a simple autonomous algorithm and a teleoperation control using an Android device sending the gyroscope generated information.
\end{abstract}

\begin{resumo} Este trabalho apresenta uma aplicação utilizando ROS, Aria e RosAria para o controle de um rôbo Pioneer 3-DX simulado em ModelSim. A navegação é feita utilizando um algoritmo de autonomia simples e pelo controle teleoperado de um dispositivo Android, que envia as informações geradas pelo giroscópio.
\end{resumo}


\section{Introdução} \label{sec:introducao}
O uso de robôs móveis autônomos é um assunto imensamente abordado na literatura de Inteligência Artificial. Segundo \cite{Ghallab:2004}, o planejamento automatizado necessita de ferramentas de processamento para que a navegação possa ser feita de maneira acessível e eficiente. Muitas vezes, essa navegação ocorre em cenários complexos e em constantes mudanças. Mesmo quando o planejamento é feito de forma extensa e minuciosa, em momentos críticos é necessário que haja intervenção humana.

Este trabalho propõe o uso de dispositivos portáteis com o sistema operacional \emph{Android} para teleoperar robôs móveis autônomos \emph{Pioneer}. Em situações sem intervação humana, o robô tomaria as ações para qual está programado, como ajustar sua trajetória e evitar obstáculos. Em momentos críticos, o operador poderia controlar o robô à distância.

O artigo está  estruturado  da seguinte forma: na Seção~\ref{sec:trabalhos} são brevemente descritos os trabalhos correlatos disponíveis na literatura. Na Seção~\ref{sec:proposta} está a descrição básica da proposta e na Seção~\ref{sec:algoritmo} são apresentados os algoritmos e implementações avaliados neste trabalho. Por fim, conclusões e direções para trabalhos futuros são descritos na Seção~\ref{sec:conclusoes}.

\section{Trabalhos correlatos} \label{sec:trabalhos}

Os campos de estudo de robôs autônomos e de navegação teleoperada são assuntos muito abordados pela literatura de Robótica. Entende-se por robô autônomo, todo robô que aceita instruções superficiais sobre suas tarefas e as executam sem a necessidade de maiores intervenções humanas~\cite{Ottoni-2000}.

\cite{Nadvornik:2014} utilizaram um robô \emph{Lego Mindstorms} teleoperado por meio de um aplicativo \emph{Android}. A comunicação é feita por um protocolo sem fio \emph{bluetooth} e baseia-se na interação por voz e por toque, sendo que a navegação ocorre com a ajuda de um sonar na parte dianteira do protótipo construído.

\cite{Selvam:2014} aplicou uma interface multimídia \emph{Android} para o reconhecimento de áreas inimigas em zonas de guerra. Utilizando uma aplicação para \emph{smartphones} fácil e intuitiva, o autor propôs controlar o robô através da interface de toque (\emph{touchscreen}).

\cite{Ko:2014} utilizaram a programação de um robô autônomo para a manutenção de estufas fazendo a pulverização de inseticidas. Este trabalho é um exemplo de como a robótica está sendo inserida em diversas áreas, tendo em vista a atual tendência da agricultura em adotar alternativas tecnológicas para diminuir o custo e o tempo de trabalho.

\cite{Chung:2004} utilizaram o algoritmo \emph{Wall Follower} e propuseram um controlador que utiliza um \emph{feedback} não-linear para ajustar a navegação do robô a uma velocidade constante e a uma distância segura de uma parede desconhecida e suave.

\section{Proposta} \label{sec:proposta}
A arquitetura do sistema robótico proposto utiliza os seguintes elementos: \emph{framework} ROS (\emph{Robotics Operating System})\footnote{ROS.org | Powering the world's robots. Disponível em: \url{http://www.ros.org}.}, 
as bibliotecas ARIA\footnote{ARIA - MobileRobots Research and Academic Customer Support. Disponível em: \url{http://robots.mobilerobots.com/wiki/ARIA}.} e RosAria\footnote{RosAria - ROS Wiki. Disponível em: \url{http://wiki.ros.org/ROSARIA}.}, o simulador MobileSim\footnote{MobileSim - MobileRobots Research and Academic Customer Support. Disponível em: \url{http://robots.mobilerobots.com/wiki/MobileSim}.}, o aplicativo \emph{ROS Android Sensor Driver}\footnote{ROS Android Sensors Driver. Disponível em: \url{https://play.google.com/store/apps/details?id=org.ros.android.sensors_driver}.} e aplicações em C++.

\begin{figure}[!htb]
    \centering
    \includegraphics[height=8.0cm]{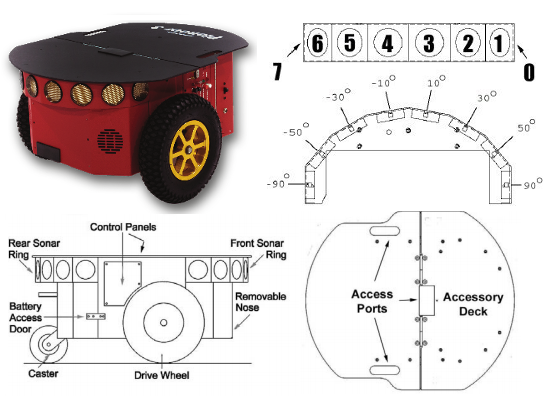}
    \caption{Robô Pioneer 3-DX}
    \label{fig:Pioneer}
\end{figure}

Para este trabalho foi utilizado o modelo \emph{Pioneer 3-DX} da \emph{MobileRobots}\footnote{Adept MobileRobots Pioneer 3-DX (P3DX) differential drive robot for research and education. Disponível em: \url{http://www.mobilerobots.com/ResearchRobots/PioneerP3DX.aspx}.}, mostrado na Figura~\ref{fig:Pioneer}. O \emph{Pioneer 3-DX} é um robô compacto com duas rodas conectadas em dois motores diferenciais. O modelo básico possui ainda odômetro e dois conjuntos de sonares (frontais e traseiros), conforme a Figura~\ref{fig:Pioneer}.

O modelo também possui diversos acessórios opcionais, tais como: telêmetro (sensores laser), \emph{bumpers} (sensores de colisão), câmera, manipuladores de 2 até 7 graus de liberdade e entre outros. A Figura~\ref{fig:PioneerAddons} ilustra um robô \emph{Pioneer 3-DX} com os seguintes acessórios:

\begin{figure}[!htb]
    \centering
    \includegraphics[height=8.0cm]{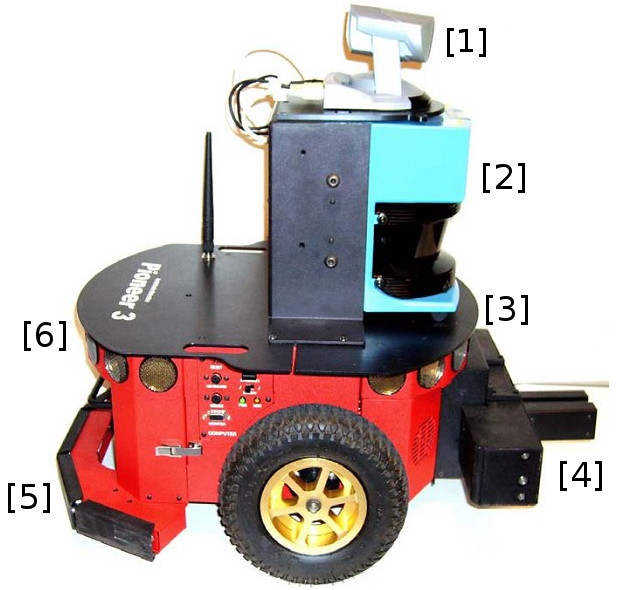}
    \caption{Robô Pioneer 3-DX com acessórios}
    \label{fig:PioneerAddons}
\end{figure}

\begin{enumerate}[label={[\arabic*]}]
    \item câmera;
    \item \emph{SICK LMS-500} (telêmetro);
    \item sensores ultrassônicos frontais;
    \item \emph{Pioneer Gripper} (manipulador com 2 DOF);
    \item \emph{bumpers} traseiros; e
    \item sensores ultrassônicos traseiros.
\end{enumerate}

Segundo informações do fabricante, o \emph{Pioneer 3-DX} é o robô móvel mais utilizado atualmente em ambientes acadêmicos e de pesquisa, com preços que iniciam em US\$ 3.995 para o modelo básico.

\section{Algoritmo e Implementação} \label{sec:algoritmo}

A aplicação foi desenvolvida em C++ e o código-fonte com instruções de execução estão disponíveis no \emph{GitHub}\footnote{Repositório tuliocasagrande/rosaria. Disponível em: \url{https://github.com/tuliocasagrande/rosaria}.}. Basicamente, o programa apresenta as seguintes funcionalidades:

\begin{enumerate}
\item Teleoperação do robô por meio da captação dos movimentos do giroscópio de um celular \emph{Android};
\item Autonomia simples, cuja navegação se baseia na detecção e emparelhamento do robô com as paredes do ambiente, também conhecido como \emph{Wall Follower}~\cite{Turennout:1992}, sendo considerado um dos mais simples algoritmos de resolução de labirintos.
\end{enumerate}

\begin{figure}[!htb]
    \centering
    \includegraphics[height=9cm]{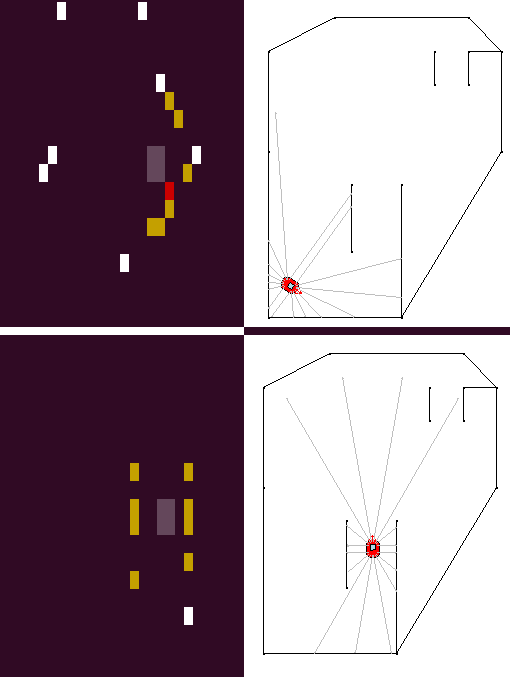}
    \caption{Sonares indicando proximidade das paredes}
    \label{fig:sonar}
\end{figure}

A aplicação também apresenta um dispositivo visual de segurança para o usuário navegar com o controle \emph{Android}. A Figura~\ref{fig:sonar} mostra em sua parte esquerda a saída do programa em execução e à direita a correspondência na simulação do robô.

\section{Conclusões} \label{sec:conclusoes}

A utilização da \emph{framework} ROS, com as bibliotecas Aria e RosAria permitem o desenvolvimento de uma interface única de comunicação, que facilita a interoperabilidade entre diversos modelos de robôs da fabricante \emph{MobileRobots}. O código proposto funciona para a maioria dos modelos da família \emph{Pioneer}, tais como: \emph{P3-DX}, \emph{P3-AT} ou \emph{AmigoBot}, com nenhuma ou poucas modificações. Além disso, mesmo se utilizados modelos de outros fabricantes, o código sofreria poucas alterações, visto que a interface proposta pelo ROS é padronizada.

As contribuições futuras consistem em refinar o algoritmo de navegação autônoma, de forma a incluir a construção de representações internas, tais como mapas e cálculos de trajetória. Além disso, a comunicação com o \emph{Android} poderia ser expandida, de forma a enviar informações multimodais para o robô, como voz ou toque. Por fim, a utilização de um robô real poderia trazer novas experiências, tais como perdas de pacotes, leituras erradas dos sensores, filtragem de ruídos e detecção de falhas.

\bibliographystyle{sbc}

\end{document}